\documentclass[10pt,twocolumn,letterpaper]{article}

\usepackage[pagenumbers]{cvpr} 

\usepackage{amsthm}
\usepackage{multirow}
\usepackage{algorithm}
\usepackage{algpseudocode}
\usepackage{xspace}
\usepackage{bm}
\newcommand{\method}{GRAFT\xspace}

\newcommand{\sd}[1]{{\scriptsize$\pm$#1}}

\DeclareMathOperator{\diag}{diag}
\newcommand{\R}{\mathbb{R}}
\newcommand{\E}{\mathbb{E}}

\theoremstyle{remark}

\AtBeginDocument{\crefname{hypothesis}{Hypothesis}{Hypotheses}\Crefname{hypothesis}{Hypothesis}{Hypotheses}\crefname{proposition}{Proposition}{Propositions}\Crefname{proposition}{Proposition}{Propositions}\crefname{algorithm}{Algorithm}{Algorithms}\Crefname{algorithm}{Algorithm}{Algorithms}}

\definecolor{cvprblue}{rgb}{0.21,0.49,0.74}
\usepackage[pagebackref,breaklinks,colorlinks,allcolors=cvprblue]{hyperref}

\def\confName{CVPR}
\def\confYear{2027}

\title{GRAFT: Graph-Regularized Prototype Fields for Realistic Test-Time Adaptation of VLMs}

\author{
Maxx Richard Rahman$^{1,2}$ \and
Wolfgang Maass$^{1,2}$ \\
\\
$^{1}$Saarland University, Germany \\
$^{2}$German Research Center for Artificial Intelligence (DFKI), Germany
}

\begin{document}
\maketitle
\begin{abstract}
Unlabeled test data can improve the zero-shot classifier of a vision-language model, yet transductive and test-time adaptation methods often fall below the zero-shot baseline when only a few vocabulary classes are present in a batch or when streams are correlated. We trace this failure to a shared design: every class owns independent parameters estimated from its own samples, so absent or rare classes are left at biased text anchors or fitted to foreign samples. We propose Graph-Regularized Anchor-displacement Field Transduction (\method), which models the adapted prototypes as text anchors plus a displacement field that is partially pooled over a semantic class graph through a Gaussian Markov random field prior, together with an explicit class-presence model. Inference is closed form and gradient free on cached features. Anchor-only shrinkage and a single global shift are the two extreme settings of the prior, absent classes are provably harmonic extensions of the present ones, and the pooling strength can be chosen by empirical Bayes. Under the realistic batch and online protocols on eleven benchmarks and four ImageNet shifts, \method improves over the zero-shot model in every setting and over anchor-based transduction where few classes are present.
\end{abstract}

\section{Introduction}
\label{sec:intro}

Vision-language models (VLMs) such as CLIP~\cite{radford2021clip} and SigLIP~\cite{zhai2023siglip} classify an image by comparing its embedding with the text embeddings of class names. The resulting zero-shot classifier is general but frozen, and a growing family of methods adapts it from the unlabeled test data itself, transductively on a batch~\cite{martin2024transductive,zanella2024transclip,kalantidis2024zlap,li2025ecalp} or online on a stream~\cite{karmanov2024tda,zhang2024dmn,zhang2024dpe,fuchs2025oga}, without training and on cached features.

These methods are evaluated on batches that contain every class of the vocabulary in comparable numbers, whereas a photo album or a user session contains a handful of the classes that the vocabulary names, and streams are temporally correlated. Under such conditions transductive methods lose tens of accuracy points with respect to the zero-shot model they are meant to improve~\cite{zanella2025stata}, and a controlled benchmark reports marginal gains, poor transfer across backbones and degraded calibration~\cite{sheng2025illusion}. The remedy proposed in~\cite{zanella2025stata} and refined in~\cite{huang2026moon} is a statistical anchor: each class prototype is shrunk towards its text embedding with a strength that grows with the number of test samples the class receives. Anchoring prevents collapse but treats every class in isolation: a class that receives no samples stays at its text anchor, although that anchor is displaced from the image distribution by the modality gap and by the domain of the test data~\cite{liang2022gap,schrodi2025two}, and a class with one or two samples is adapted from those samples alone. The problem addressed here is therefore to adapt \emph{all} prototypes of a large vocabulary from an unlabeled batch in which most classes are absent or rare, without collapsing on the absent ones and without discarding what the present ones reveal about them.

\begin{figure}[t]
  \centering
  \includegraphics[width=\linewidth]{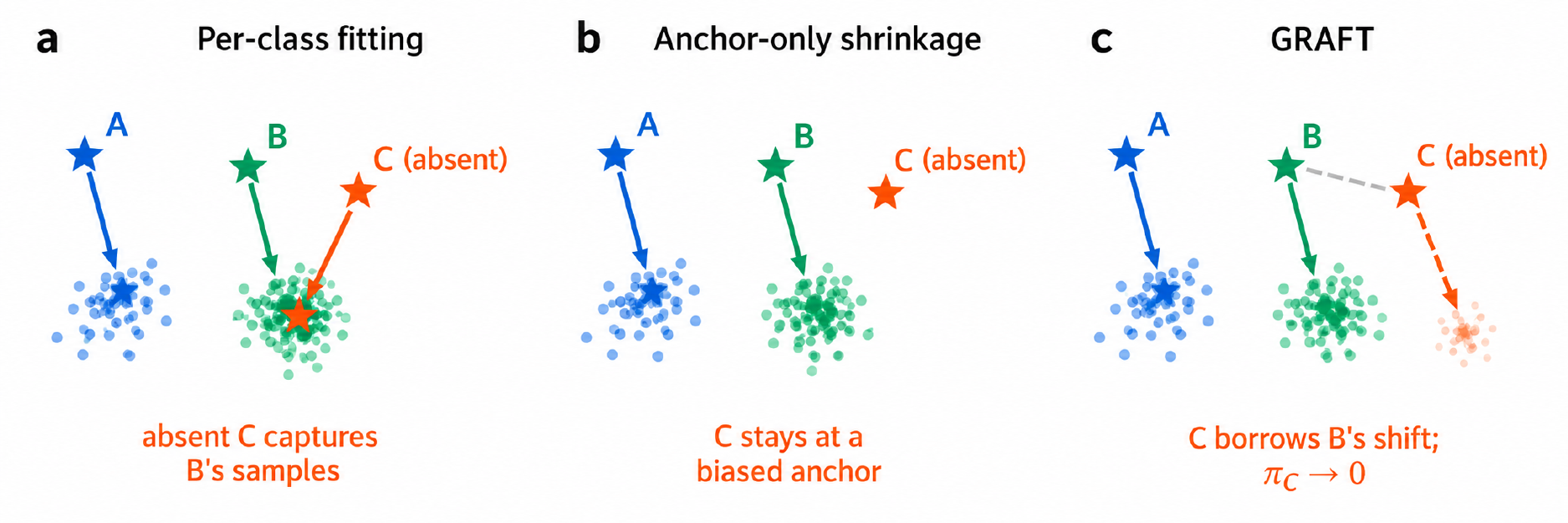}
  \caption{Per-class adaptation on a realistic batch (conceptual illustration). Classes A and B are present, C is absent, and all images are shifted from their anchors. (a) Per-class fitting lets C capture samples of B. (b) Anchor-only shrinkage leaves C at its biased anchor. (c) \method lets C inherit B's displacement and removes it through the presence model.}
  \label{fig:concept}
\end{figure}

The displacement between the text anchor of a class and the centroid of its images contains a component shared by all classes, the modality and domain offset, and a component that varies smoothly with the semantics of the class. \method turns this structure into a prior. The adapted prototypes are written as $\mu_k = t_k + d_k$ and the displacement field $D = [d_1;\dots;d_K]$ receives a Gaussian Markov random field prior whose precision is $\alpha I + \lambda L_T$, where $L_T$ is the Laplacian of a nearest-neighbour graph over the text embeddings. The $\alpha$ term reproduces the anchor-only shrinkage of~\cite{zanella2025stata,huang2026moon}. The $\lambda$ term couples the classes, so that the displacement of a well-observed class propagates to its semantic neighbours. The maximum a posteriori field has a closed form: a graph filter that returns the empirical displacement for well-observed classes and the harmonic extension of the neighbours' displacements for absent ones (\cref{fig:concept}c). Classes that receive no samples are additionally removed from the competition by a Dirichlet-regularised estimate of the class proportions, which treats the vocabulary as a label-shift problem~\cite{saerens2002adjusting,garg2020unified}. All updates are closed form on cached features, and the prior strengths can be selected by maximising the marginal likelihood of the unlabeled batch, which removes the dataset-specific tuning criticised in~\cite{sheng2025illusion}.

\paragraph{Contributions.}
\begin{itemize}
  \item We formulate transductive VLM adaptation as the estimation of a displacement field over the class vocabulary with a graph-structured prior, and derive closed-form block-coordinate inference for batches and streams (\cref{sec:method}); anchor-only shrinkage and a single global shift are its two extreme settings, and absent classes receive the harmonic extension of the present-class displacements.
  \item We combine the field with an explicit class-presence model and an empirical Bayes rule for the prior strength, so that the vocabulary is treated as a label-shift problem and no pooling strength has to be tuned per dataset.
  \item We evaluate \method under the realistic batch and online protocols on eleven benchmarks and four ImageNet shifts, with architecture-matched controls that isolate the graph coupling and the presence model.
\end{itemize}

\section{Related Work}
\label{sec:related}

\paragraph{Unlabeled adaptation of vision-language models.}
Transductive methods classify an unlabeled batch jointly: TransCLIP~\cite{zanella2024transclip} fits a Gaussian mixture regularised by a KL term to the zero-shot posterior and a sample-graph Laplacian, EM-Dirichlet~\cite{martin2024transductive} works on the simplex, ZLaP~\cite{kalantidis2024zlap} and ECALP~\cite{li2025ecalp} propagate labels over image and prompt graphs, and LATA~\cite{bozorgtabar2026lata} and Conf-OT~\cite{silva2025confot} target conformal prediction. Online methods adapt over a stream with feature caches~\cite{karmanov2024tda,zhang2024dmn,zhang2024boostadapter}, evolving prototypes~\cite{zhang2024dpe}, per-class Gaussians~\cite{fuchs2025oga} or online label distributions~\cite{qian2024onzeta,han2026lsa}; episodic methods act on single images~\cite{shu2022tpt,zanella2024mta,farina2024zero,sui2025tps}. The realistic protocols with a variable number of effective classes and correlated streams were introduced in~\cite{zanella2025stata}, which showed that these methods collapse under them and proposed StatA, a shrinkage of each class's Gaussian parameters towards its text embedding with strength $n_k/(n_k+\alpha)$. MOON~\cite{huang2026moon} derives this shrinkage from penalised likelihood on the hypersphere and makes it confidence dependent. Both anchor every class to its own text embedding and use no information from other classes; their mean update is the $\lambda=0$ case of \method. A unified benchmark further reports marginal gains, poor transfer to SigLIP and worse calibration~\cite{sheng2025illusion}.

\paragraph{Structure over classes and label shift.}
Class graphs propagate classifier weights to unseen classes in zero-shot learning~\cite{kampffmeyer2019dgp}, propagate prototypes along a category hierarchy~\cite{liu2019ppn} and select correlated classes in CLIP test-time adaptation~\cite{huang2025cosmic}. Laplacian regularisation over samples is standard in transductive few-shot learning~\cite{ziko2020laplacianshot,zhou2004learning}. We are not aware of prior work that regularises the \emph{displacement} of VLM prototypes from their text anchors with a Gaussian Markov random field over the class graph. The modality gap literature~\cite{liang2022gap,schrodi2025two} supports the premise that much of the displacement is shared. Adjusting a classifier to new class priors from unlabeled data is classical~\cite{saerens2002adjusting,garg2020unified}; Frolic~\cite{zhu2024frolic} corrects the label bias of CLIP without labels, and open-vocabulary classifiers degrade as distractor classes are added~\cite{chen2026vocab}.

\section{Preliminaries}
\label{sec:prelim}

\paragraph{Zero-shot classification.}
A VLM provides an image encoder $f_v$ and a text encoder $f_t$ mapping into a common space $\R^d$. For a vocabulary of $K$ class names $\{c_k\}_{k=1}^K$, prompt ensembling yields unit-norm text anchors $t_k = f_t(c_k)/\|f_t(c_k)\|$, collected in $T\in\R^{K\times d}$. A test image $I_i$ gives $x_i = f_v(I_i)/\|f_v(I_i)\|$, collected in $X\in\R^{N\times d}$. The zero-shot posterior is
\begin{equation}
  \hat{y}_{ik} = \frac{\exp(\tau\, x_i^{\top} t_k)}{\sum_{l}\exp(\tau\, x_i^{\top} t_l)},
  \label{eq:zs}
\end{equation}
with logit scale $\tau$ (100 for CLIP). We write $\Delta^K$ for the probability simplex and $\mathbf{1}$ for the all-ones vector.

\paragraph{Unlabeled adaptation.}
Given $X$ and $T$ only, the goal is to output posteriors $z_i\in\Delta^K$ more accurate and at least as well calibrated as $\hat{y}_i$, for a batch processed at once or for a stream of batches predicted as they arrive. Following~\cite{zanella2025stata}, a batch contains $K_{\mathrm{eff}}\le K$ effective classes, often far fewer than the vocabulary, and streams are correlated.

\paragraph{Mixture-model transduction and its anchors.}
TransCLIP~\cite{zanella2024transclip} and StatA~\cite{zanella2025stata} model $x_i$ with a $K$-component Gaussian mixture and minimise, over assignments $z$ and parameters, a penalised negative log-likelihood with a KL anchor to $\hat{y}$ and a Laplacian term over a nearest-neighbour graph of the samples. StatA additionally shrinks the mean of class $k$ towards its anchor,
\begin{equation}
  \mu_k = \beta_k\,\bar{x}_k + (1-\beta_k)\,t_k, \qquad \beta_k = \frac{n_k}{n_k+\alpha},
  \label{eq:stata}
\end{equation}
with $n_k = \sum_i z_{ik}$ the soft count and $\bar{x}_k$ the responsibility-weighted mean. \Cref{eq:stata} is the posterior mean of $\mu_k$ under an independent Gaussian prior centred at $t_k$: a class with $n_k=0$ keeps $\mu_k = t_k$ even when every present class has moved by a common offset, a class with small $n_k$ is estimated from its own few samples only, and the mixture weights are implicitly uniform, so absent classes keep competing for samples. \method addresses the first two points by coupling the classes through a graph prior and the third by estimating the class proportions.

\section{Method}
\label{sec:method}

\begin{figure*}[t]
  \centering
  \includegraphics[width=0.95\textwidth]{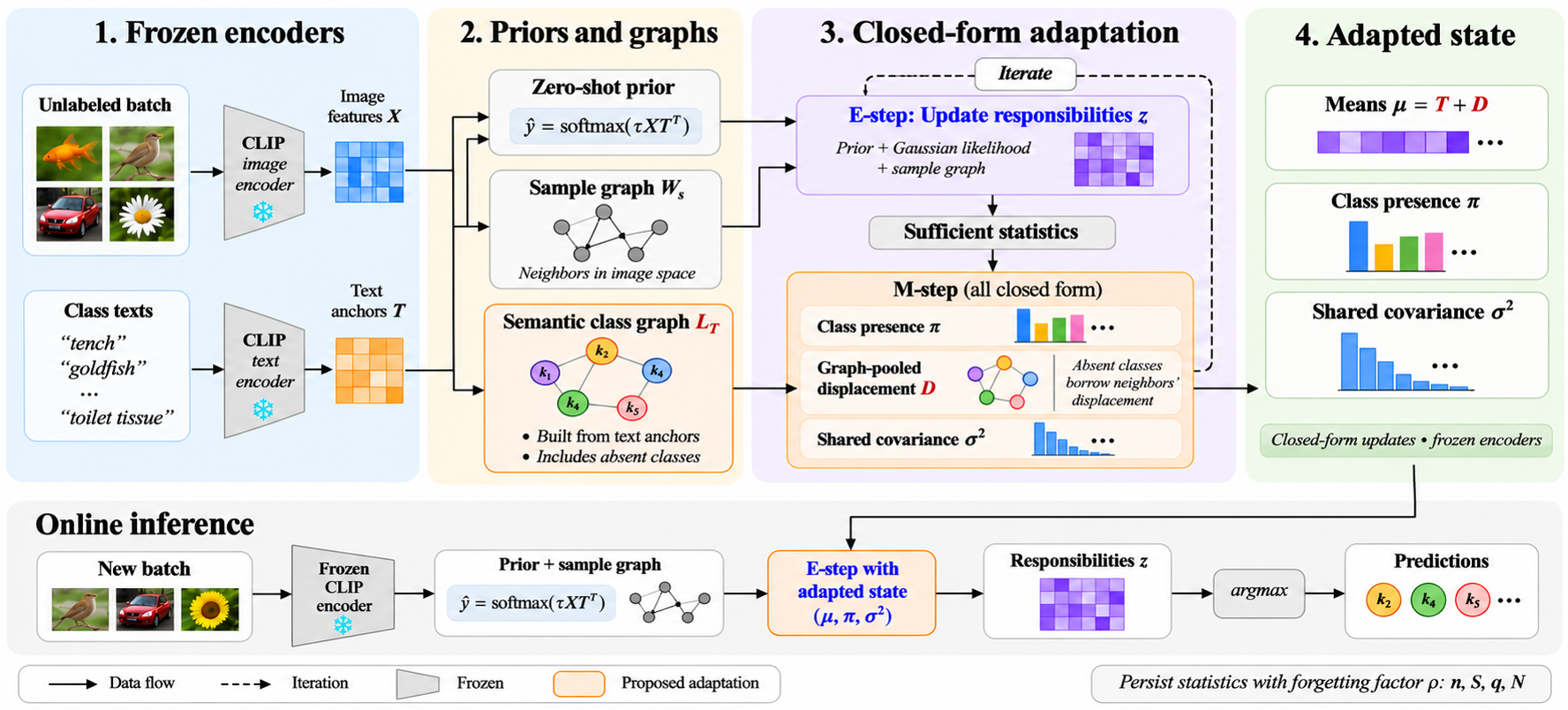}
  \caption{Overview of \method. Frozen encoders give features $X$ and anchors $T$, which define the zero-shot prior $\hat{y}$, the sample graph $W_s$ (from $X$, images to images) and the semantic class graph $L_T$ (from $T$, classes to classes). Closed-form updates alternate the E-step over $z$, using $\hat{y}$ and $W_s$, with M-steps for the presence $\pi$, the displacement field $D$ pooled over $L_T$ and the shared covariance; $\mu = T + D$ are the adapted prototypes and $z$ the prediction.}
  \label{fig:framework}
\end{figure*}

\method keeps the generative view of transduction~\cite{zanella2024transclip,zanella2025stata} but changes the object that is adapted. Instead of $K$ independent prototypes it estimates a \emph{displacement field} $D\in\R^{K\times d}$ anchored at the text embeddings, $\mu_k = t_k + d_k$, under a prior that couples semantically related classes, together with the class proportions $\pi\in\Delta^K$ and a shared covariance. \Cref{fig:framework} shows the components: the frozen encoders with the zero-shot prior of \cref{eq:zs} and the two graphs (\cref{sec:graphs}), the likelihood and the graph-regularised field (\cref{sec:field}), the class-presence model (\cref{sec:presence}), the objective (\cref{sec:objective}), the closed-form inference (\cref{sec:inference}) and the empirical Bayes rule for the prior strength (\cref{sec:eb}).

\subsection{Sample graph and semantic class graph}
\label{sec:graphs}
Two graphs enter the model, and they are built from different sources. The \emph{sample graph} $W_s\in\R^{N\times N}$, inherited from~\cite{zanella2024transclip,zanella2025stata}, is data driven: it connects each test image to its $m_s$ nearest neighbours in feature space with weight $(W_s)_{ii'} = x_i^{\top}x_{i'}$ and encodes the manifold assumption that neighbouring images share a label. It is recomputed for every batch and says nothing about classes that have no image in the batch.

The \emph{semantic class graph}, introduced in this work, is built from the vocabulary only. Its nodes are the $K$ classes and its edges are derived from the text anchors $T$ produced by the frozen text encoder, so that it depends on the class names and the prompts, not on the images, the batch or the responsibilities. Let $S = TT^{\top}$ be the cosine similarity between anchors. Class $k$ is connected to the $m$ classes with the largest $S_{kl}$, $l\neq k$, with the heat-kernel weight
\begin{equation}
  w_{kl} = \exp\big((S_{kl}-1)/\rho\big)\in(0,1],
  \label{eq:classgraph}
\end{equation}
with bandwidth $\rho$. The neighbourhoods are symmetrised by union, $W = \max(W, W^{\top})$, and the normalised Laplacian is $L_T = I - \mathrm{D}^{-1/2} W \mathrm{D}^{-1/2}$ with $\mathrm{D} = \diag(W\mathbf{1})$. $L_T\in\R^{K\times K}$ is positive semi-definite with spectrum in $[0,2]$ and is computed once per vocabulary. Because it is built from text semantics, every class has neighbours, including those that receive no sample in a batch, so the field below is defined on the whole vocabulary. The graph expresses the hypothesis, that classes with similar names are displaced from their anchors in similar directions.

\subsection{Graph-regularised displacement field}
\label{sec:field}
\paragraph{Likelihood.}
Each class is a Gaussian with mean $\mu_k = t_k + d_k$ and a diagonal covariance $\Sigma=\diag(\sigma^2)$ shared across classes. As in TransCLIP, the likelihood is tempered by $\kappa$, which acts as a covariance inflation and keeps the responsibilities soft on $d$-dimensional features:
\begin{equation}
  \ell_{ik}(D,\sigma) = -\frac{1}{2\kappa}\Big[\sum_{j=1}^{d}\frac{(x_{ij}-t_{kj}-d_{kj})^2}{\sigma_j^2} + \sum_{j=1}^{d}\log\sigma_j^2\Big].
  \label{eq:lik}
\end{equation}
Given responsibilities $z$, the shared variance has the closed-form M-step $\sigma_j^2 = \tfrac{1}{N}\sum_{i,k} z_{ik}(x_{ij}-\mu_{kj})^2$.

\paragraph{How the graph defines the field.}
The displacement field $D\in\R^{K\times d}$ assigns to every class $k$ of the vocabulary the vector $d_k$ by which its prototype departs from its text anchor. The class graph enters through a Gaussian Markov random field prior~\cite{rue2005gmrf} on this field, which factorises over feature dimensions,
\begin{equation}
  p(D) \propto \exp\Big(-\tfrac{1}{2}\sum_{j=1}^{d} d_{:j}^{\top} Q\, d_{:j}\Big),\qquad
  Q = \frac{\alpha I_K + \lambda L_T}{\kappa\,\bar{\sigma}^2},
  \label{eq:prior}
\end{equation}
with $\bar{\sigma}^2 = \tfrac{1}{d}\sum_j\sigma_j^2$. The two terms of the precision $Q$ have distinct roles. The term $\alpha I_K$ pulls every displacement towards zero, i.e., every prototype towards its own text anchor; this is the anchoring of StatA and MOON. The term $\lambda L_T$ is the coupling introduced here: it penalises the Dirichlet energy $\sum_{k\sim l} w_{kl}\|d_k/\sqrt{\mathrm{D}_k} - d_l/\sqrt{\mathrm{D}_l}\|^2$, so that displacements of classes joined by an edge are asked to agree, with a strength given by the semantic weight $w_{kl}$. The conditional of the GMRF makes the role of the graph explicit: under \cref{eq:prior}, the prior mean of $d_k$ given the displacements of all other classes is
\begin{equation}
  \E[d_k \,|\, d_{-k}] = \frac{\lambda}{\alpha+\lambda}\sum_{l\sim k}\frac{w_{kl}}{\sqrt{\mathrm{D}_k\mathrm{D}_l}}\, d_l ,
  \label{eq:gmrf-cond}
\end{equation}
a weighted average of the neighbours' displacements on the class graph, shrunk towards zero by the anchor. A class therefore inherits, a priori, the displacement of the classes whose names are closest to its own. Dividing by $\kappa\bar{\sigma}^2$ expresses $\alpha$ and $\lambda$ in units of sample counts.

\begin{figure}[t]
  \centering
  \includegraphics[width=0.82\linewidth]{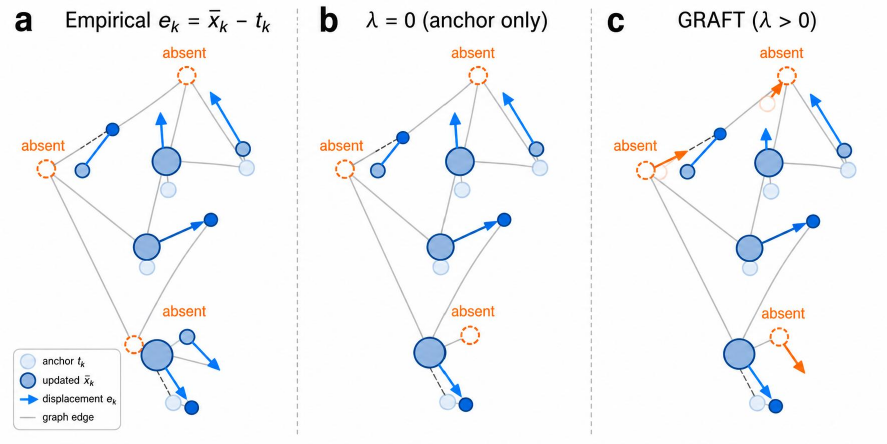}
  \caption{Graph-pooled displacement field on a toy class graph (node size: $n_k$). (a) Empirical displacements. (b) Independent shrinkage ($\lambda=0$) leaves absent classes at their anchors. (c) \cref{eq:dsolve} transfers displacements along the graph.}
  \label{fig:module}
\end{figure}

\paragraph{How the field is regularised and updated.}
Three forces act on every displacement. The data term of the likelihood pulls $d_k$ towards the empirical displacement $e_k = \bar{x}_k - t_k$ of the images currently assigned to class $k$, with weight equal to the soft count $n_k=\sum_i z_{ik}$ (here $\bar{x}_k = \sum_i z_{ik}x_i/n_k$, and $e_k=0$ when $n_k=0$); the anchor term pulls it towards zero with weight $\alpha$; and the coupling term pulls it towards the neighbour average of \cref{eq:gmrf-cond} with weight $\lambda$. Writing $N_z=\diag(n_1,\dots,n_K)$, the maximum a posteriori field given the responsibilities balances the three and is obtained for every feature dimension $j$ from a $K\times K$ linear system,
\begin{equation}
  d_{:j} = \big(N_z + s_j(\alpha I_K + \lambda L_T)\big)^{-1} N_z\, e_{:j}, \qquad s_j = \sigma_j^2/\bar{\sigma}^2 .
  \label{eq:dsolve}
\end{equation}
\Cref{eq:dsolve} is a graph filter applied to the empirical displacements (\cref{fig:module}). Writing $\tilde{A} = \mathrm{D}^{-1/2}W\mathrm{D}^{-1/2}$ for the normalised adjacency and $M = N_z + s_j(\alpha+\lambda)I_K$ for the diagonal part, it expands as the convergent series
\begin{equation}
  d_{:j} = \sum_{t\ge 0}\big(s_j\lambda\, M^{-1}\tilde{A}\big)^{t} M^{-1} N_z\, e_{:j},
  \label{eq:diffusion}
\end{equation}
in which the evidence-weighted empirical displacements $N_z e_{:j}$ are diffused along the class graph: the term $t$ carries the displacement of present classes to classes $t$ edges away, attenuated at every hop by $s_j\lambda/(n_l + s_j(\alpha+\lambda))$. Hence a class with $n_k \gg \alpha+\lambda$ keeps its empirical displacement, a class with few samples mixes its own noisy estimate with its neighbours' displacements in proportion to $n_k$ against $\lambda$, and a class with no sample receives only what diffuses from its neighbours. The field is re-estimated at every outer iteration: the responsibilities $z$ determine $n_k$ and $e_k$, \cref{eq:dsolve} gives $D$, the prototypes $\mu = T + D$ enter the likelihood of \cref{eq:lik}, and the next E-step updates $z$ (\cref{sec:inference}); the $d$ systems are solved jointly by preconditioned conjugate gradients.

\paragraph{Special cases.}
With $\lambda=0$, \cref{eq:dsolve} reduces per dimension to $d_{kj} = \frac{n_k}{n_k+\alpha s_j}\,e_{kj}$, the count-based anchor shrinkage of StatA (\cref{eq:stata}) and MOON~\cite{huang2026moon}, which are therefore the graph-free member of the family. With $\alpha=0$, $L_T$ the Laplacian of the complete graph and $\lambda\to\infty$, every class receives the count-weighted mean displacement $d = \sum_k n_k e_k / \sum_k n_k$, a single global shift of all prototypes. \method interpolates between these extremes along the semantic neighbourhoods.

\paragraph{Absent classes.}
Let $A=\{k: n_k = 0\}$ be the absent classes, $P$ their complement and $Q_\lambda = \alpha I + \lambda L_T$. Since the rows of \cref{eq:dsolve} in $A$ have $n_k=0$ and $e_k=0$, the solution satisfies $Q_{\lambda,AA}\,D_A + Q_{\lambda,AP}\,D_P = 0$, so
\begin{equation}
  D_A = -Q_{\lambda,AA}^{-1}Q_{\lambda,AP}\,D_P
  \label{eq:harmonic}
\end{equation}
is the harmonic extension of the present-class field, and $D_P$ solves \cref{eq:dsolve} on $P$ with $Q_\lambda$ replaced by its Kron reduction $Q_\lambda/Q_{\lambda,AA}$~\cite{dorfler2013kron}: absent classes affect the present-class prototypes only through the effective graph, never through data (\cref{fig:concept}c).

\paragraph{Posterior variance.}
Given $z$, the posterior covariance of $d_{:j}$ is $\kappa\sigma_j^2\,(N_z + s_j Q_\lambda)^{-1}$; since $\lambda L_T\succeq 0$, $\mathrm{Var}(d_{kj}) \le \kappa\sigma_j^2/(n_k + s_j\alpha)$, the variance under independent shrinkage. Proofs are in the supplementary material.

\paragraph{Smoothness hypothesis.}
None of these statements guarantees higher accuracy than zero-shot inference. The benefit of the coupling rests on a testable hypothesis, class-level smoothness (H1): the oracle displacement field $e^{\star}_k = \E[x\,|\,y=k] - t_k$ has a Dirichlet energy on $L_T$ substantially smaller than on a degree-matched random graph. If H1 fails, \method reduces to anchor-only shrinkage.

\subsection{Class presence as label shift}
\label{sec:presence}
The mixture weights $\pi$ are free parameters with a symmetric Dirichlet prior $\mathrm{Dir}(c+1)$. Their M-step is the smoothed count
\begin{equation}
  \pi_k = \frac{n_k + c}{N + Kc},
  \label{eq:pi}
\end{equation}
and they enter the E-step as $\log\pi_k$, which is the classical label-shift correction of a calibrated classifier~\cite{saerens2002adjusting}: the zero-shot posterior implicitly assumes uniform class priors, and \cref{eq:pi} replaces them by the estimate obtained from the batch. For a batch with $K_{\mathrm{eff}}\ll K$ classes, absent classes receive $\pi_k\approx c/(N+Kc)$ while present ones receive $n_k/(N+Kc)$: with $c=\tfrac{1}{2}$, a present class with sixteen samples is more than thirty times more likely a priori than an absent one in a batch of 64 over 1000 classes. The pseudo-count keeps $\log\pi_k$ finite and lets a class absent from earlier online batches re-enter the stream.

\subsection{Objective}
\label{sec:objective}
Collecting the terms, \method minimises over $(z, D, \sigma, \pi)$
\begin{align}
  J ={}& \sum_{i,k} z_{ik}\big[\log z_{ik} - \log\pi_k - \lambda_y\log\hat{y}_{ik} - \ell_{ik}(D,\sigma)\big] \nonumber\\
   &- \frac{\lambda_L}{2m_s}\sum_{i,i'} (W_s)_{ii'}\, z_i^{\top} z_{i'}
    + \frac{1}{2}\sum_{j} d_{:j}^{\top} Q\, d_{:j}
    - c\sum_k \log\pi_k ,
  \label{eq:objective}
\end{align}
where the first line is the variational free energy of the mixture with the text prior $\hat{y}^{\lambda_y}$ of~\cite{zanella2024transclip}, and the second line contains the inherited sample-graph term, the field prior of \cref{eq:prior} and the presence prior. With $\lambda = 0$ and without the $\pi$ terms this is the StatA objective up to its per-class covariances; with $\alpha=\lambda_L=0$ in addition, the unanchored mixture.

\subsection{Inference}
\label{sec:inference}
The objective is minimised by block-coordinate descent over four blocks (pseudocode in the supplementary material). The E-step is the mean-field update with the Laplacian term linearised at the current $z$, a majorise-minimise step~\cite{hunter2004mm,ziko2020laplacianshot,zanella2024transclip}:
\begin{equation}
  z_{ik} \propto \pi_k\,\hat{y}_{ik}^{\lambda_y}\exp\Big(\ell_{ik} + \frac{\lambda_L}{2m_s}\big[(W_s + W_s^{\top})z\big]_{ik}\Big),
  \label{eq:estep}
\end{equation}
iterated a few times. The M-steps are the presence update of \cref{eq:pi}, the field of \cref{eq:dsolve} and the shared variance of \cref{sec:field}. Each block update is the exact minimiser of \cref{eq:objective} over its block when $\lambda_L=0$, so $J$ decreases monotonically and converges; with $\lambda_L>0$ the E-step minimises a majoriser of $J$, as in~\cite{zanella2024transclip}. All quantities depend on the data only through the sufficient statistics $n = z^{\top}\mathbf{1}$, $S = z^{\top}X$ and $q = \sum_i x_i^{\odot 2}$, which makes the online variant immediate: the statistics are accumulated over the stream with an optional forgetting factor $\varrho$, and the same M-steps are applied after every batch. The prediction for a batch is its posterior $z$ after adaptation, as in the transductive protocol of~\cite{zanella2025stata}; a strictly inductive variant predicts before the statistics are updated. Per outer iteration the E-step costs $O(NKd + N m_s K)$ and the field $O(n_{\mathrm{cg}}\,K d\,(1+m))$ with conjugate gradients. No backward pass is executed and the VLM is untouched after feature extraction.

\subsection{Empirical Bayes selection of the pooling strength}
\label{sec:eb}
Fixed $(\alpha,\lambda)$ are used for all datasets in the main protocol, but the model also admits a label-free rule: under the model, the empirical displacements of the present classes are Gaussian with a covariance that depends on $(\alpha,\lambda)$ through the Schur complement of $Q$ onto the present classes, and their marginal likelihood is maximised over a small logarithmic grid at every M-step whenever at least three classes are present. The derivation is given in the supplementary material.

\section{Experiments}
\label{sec:experiments}

\subsection{Datasets}
We follow the eleven-dataset suite of TransCLIP and StatA~\cite{zanella2024transclip,zanella2025stata} with the CoOp test splits~\cite{zhou2022coop}: ImageNet~\cite{deng2009imagenet}, SUN397~\cite{xiao2010sun}, FGVC-Aircraft~\cite{maji2013aircraft}, EuroSAT~\cite{helber2019eurosat}, Stanford Cars~\cite{krause2013cars}, Food101~\cite{bossard2014food}, Oxford Pets~\cite{parkhi2012pets}, Flowers102~\cite{nilsback2008flowers}, Caltech101~\cite{feifei2004caltech}, DTD~\cite{cimpoi2014dtd} and UCF101~\cite{soomro2012ucf101}, plus the ImageNet distribution shifts ImageNet-V2~\cite{recht2019imagenetv2}, ImageNet-Sketch~\cite{wang2019sketch}, ImageNet-A~\cite{hendrycks2021imageneta} and ImageNet-R~\cite{hendrycks2021imagenetr}. Class names and prompt templates are those of CoOp and StatA. Features are extracted once per backbone and cached. No training split is used. Batch adaptation follows the seven settings of~\cite{zanella2025stata}: a task draws $K_{\mathrm{eff}}$ classes uniformly from $[1,4]$, $[2,10]$ or $[5,25]$ (very low, low, medium) with $B{=}64$ images, from $[5,25]$, $[25,50]$ or $[50,100]$ (medium, high, very high) with $B{=}1000$, or uses the full test set, with 1000 tasks per dataset. Online adaptation streams batches of 128 whose class composition follows Dirichlet($\gamma$) slot proportions, $\gamma\in\{0.1,0.01,0.001\}$ (low to high correlation) or one class after another (separate), with 100 runs per dataset. The vocabulary sizes range from 10 (EuroSAT) to 1000 (ImageNet) and the test sets from 1,692 (DTD) to 50,000 images (ImageNet).

\subsection{Baselines}
We compare with zero-shot CLIP, the episodic MTA~\cite{zanella2024mta}, the transductive EM-Dirichlet~\cite{martin2024transductive}, ZLaP~\cite{kalantidis2024zlap}, TransCLIP~\cite{zanella2024transclip}, StatA~\cite{zanella2025stata} and MOON~\cite{huang2026moon}, and the online TENT~\cite{wang2021tent}, TDA~\cite{karmanov2024tda}, DMN~\cite{zhang2024dmn} and OGA~\cite{fuchs2025oga}, all run with default hyper-parameters through the StatA harness into which \method is plugged as a drop-in solver, so that samplers, features and prompts are identical. To attribute gains to the proposed mechanism rather than to the likelihood or the inherited terms, four architecture-matched controls share our implementation: anchor-only shrinkage ($\lambda{=}0$, the StatA/MOON mechanism in our likelihood), complete pooling (a single global shift: $\alpha{=}0$, complete graph, $\lambda\to\infty$), presence only ($\pi$ without displacement) and the unanchored mixture ($\alpha{=}\lambda{=}0$, no $\pi$).

\subsection{Implementation details}
Inherited constants follow TransCLIP and StatA and are never tuned: $\kappa{=}50$, $\lambda_y{=}1$, $\lambda_L{=}1$, $m_s{=}3$, $T_{\max}{=}10$ outer iterations with 3 inner E-step sweeps. The new constants are fixed once for all datasets and backbones: $\alpha{=}1$ (one pseudo-count, as in StatA), $\lambda{=}4$, $m{=}5$ neighbours, $\rho{=}0.05$, $c{=}\tfrac12$. The linear systems are solved exactly for $K\le256$ and by 30 preconditioned conjugate-gradient iterations otherwise. Backbones: CLIP ViT-B/16 (main), ViT-L/14 and RN50~\cite{radford2021clip}, OpenCLIP~\cite{cherti2023openclip} ViT-B/16 trained on DataComp~\cite{gadre2023datacomp}, and SigLIP ViT-B/16~\cite{zhai2023siglip}, the backbone on which~\cite{sheng2025illusion} found most TTA methods to fail. All runs use a single GPU. Seeds and task indices are fixed and shared across methods. We report top-1 accuracy averaged over tasks and datasets, the worst-decile accuracy, expected calibration error and time per task in supplementary material.

\section{Results}
\label{sec:results}

\subsection{Realistic batches}
\begin{figure}[t]
  \centering
  \includegraphics[width=0.72\linewidth]{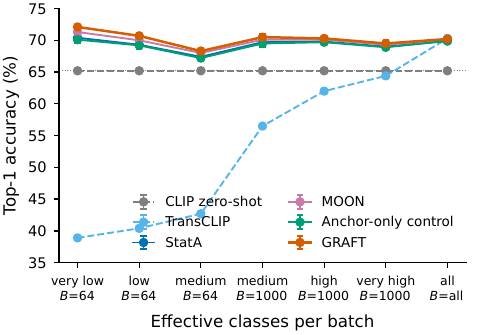}
  \caption{Top-1 accuracy versus the number of effective classes per batch (eleven-dataset mean, error bars: standard deviation over tasks); the dotted line is the zero-shot model.}
  \label{fig:keff}
\end{figure}

\Cref{tab:main} and \cref{fig:keff} compare \method with the state-of-the-art methods and the architecture-matched controls across the seven batch settings. First, the unanchored transductive methods collapse as the number of effective classes shrinks: TransCLIP falls from 70.3 with all classes to 38.9 in the very-low setting and ZLaP from 66.4 to 27.5, whereas EM-Dirichlet, the strongest method at $B{=}64$, collapses in the opposite direction, to 29.5 on the full set. The unanchored control of our implementation reproduces the first pattern (42.4 to 45.4 at $B{=}64$). Second, anchoring removes the collapse but adds little beyond it: StatA, MOON and the anchor-only control lie within 1.2 points of each other in every setting, so the count-dependent shrinkage they share is what prevents collapse. Third, pooling the field across classes adds a margin on top of anchoring that is largest where absent and rare classes dominate and vanishes where every class is well observed: \method improves over the anchor-only control by 1.9, 1.6 and 1.1 points in the very-low, low and medium settings with $B{=}64$, by 0.9, 0.7 and 0.5 points with $B{=}1000$, and by 0.3 points on the full set, where \cref{sec:field} predicts that the two mechanisms coincide; against MOON the margins are 0.7 to 0.1 points. The presence model alone reaches 67.4 in the very-low setting, about a third of the gain over zero-shot, and the complete-graph control, a single shift of all prototypes, stays below the anchor-only control in every setting, so the gain comes from pooling along the semantic neighbourhoods rather than from a global shift. Empirical Bayes selection of $(\alpha,\lambda)$ matches the fixed strengths within 0.1 point, and standard deviations over tasks are 0.2 to 0.4 points for the anchored methods, well below these margins.

\begin{table*}[t]
  \centering\footnotesize
  \caption{Batch adaptation with a variable number of effective classes: top-1 accuracy (\%) averaged over 1000 tasks and eleven datasets, CLIP ViT-B/16.}
  \label{tab:main}
  \setlength{\tabcolsep}{4.5pt}
  \begin{tabular}{lccccccc}
    \toprule
    \textbf{Method} & \textbf{Very low} & \textbf{Low} & \textbf{Medium} & \textbf{Medium} & \textbf{High} & \textbf{Very high} & \textbf{All}\\
     & $B{=}64$ & $B{=}64$ & $B{=}64$ & $B{=}1000$ & $B{=}1000$ & $B{=}1000$ & full\\
    \midrule
    CLIP zero-shot~\cite{radford2021clip} & 65.2\sd{0.6} & 65.2\sd{0.5} & 65.2\sd{0.4} & 65.2\sd{0.2} & 65.2\sd{0.2} & 65.2\sd{0.1} & 65.2\\
    MTA~\cite{zanella2024mta} & 66.6\sd{0.6} & 66.6\sd{0.5} & 66.6\sd{0.4} & 66.6\sd{0.2} & 66.6\sd{0.2} & 66.6\sd{0.1} & 66.6\\
    EM-Dirichlet~\cite{martin2024transductive} & 68.5\sd{0.7} & 70.3\sd{0.6} & 67.5\sd{0.5} & 64.4\sd{0.4} & 45.3\sd{0.6} & 33.6\sd{0.7} & 29.5\\
    ZLaP~\cite{kalantidis2024zlap} & 27.5\sd{1.1} & 35.2\sd{0.9} & 44.7\sd{0.8} & 41.5\sd{0.5} & 52.2\sd{0.4} & 58.4\sd{0.3} & 66.4\\
    TransCLIP~\cite{zanella2024transclip} & 38.9\sd{1.0} & 40.4\sd{0.9} & 42.7\sd{0.8} & 56.5\sd{0.4} & 62.0\sd{0.3} & 64.4\sd{0.2} & \textbf{70.3}\\
    StatA~\cite{zanella2025stata} & 70.4\sd{0.4} & 69.3\sd{0.3} & 67.4\sd{0.3} & 69.7\sd{0.2} & 69.8\sd{0.2} & 69.0\sd{0.2} & 69.9\\
    MOON~\cite{huang2026moon} & 71.3\sd{0.4} & 70.0\sd{0.3} & 67.9\sd{0.3} & 70.1\sd{0.2} & 70.1\sd{0.2} & 69.3\sd{0.2} & 70.1\\
    \midrule
    Unanchored mixture ($\alpha{=}\lambda{=}0$, no $\pi$) & 42.4\sd{0.6} & 43.3\sd{0.5} & 45.4\sd{0.5} & 57.7\sd{0.3} & 62.8\sd{0.3} & 64.5\sd{0.2} & 70.3\\
    Presence only ($\pi$, no field) & 67.4\sd{0.5} & 66.7\sd{0.4} & 65.3\sd{0.4} & 66.9\sd{0.2} & 66.6\sd{0.2} & 65.7\sd{0.2} & 66.0\\
    Anchor-only shrinkage ($\lambda{=}0$) & 70.1\sd{0.4} & 69.1\sd{0.3} & 67.3\sd{0.3} & 69.6\sd{0.2} & 69.7\sd{0.2} & 69.0\sd{0.2} & 69.9\\
    Complete pooling (global shift) & 69.6\sd{0.4} & 68.8\sd{0.3} & 66.8\sd{0.3} & 68.8\sd{0.2} & 68.6\sd{0.2} & 67.6\sd{0.2} & 68.3\\
    \method (fixed $\alpha,\lambda$) & \underline{72.0}\sd{0.4} & \underline{70.7}\sd{0.3} & \underline{68.4}\sd{0.3} & \underline{70.5}\sd{0.2} & \underline{70.4}\sd{0.2} & \underline{69.5}\sd{0.2} & \underline{70.2}\\
    \method + empirical Bayes & \textbf{72.1}\sd{0.4} & \textbf{70.8}\sd{0.3} & \textbf{68.5}\sd{0.3} & \textbf{70.6}\sd{0.2} & \textbf{70.5}\sd{0.2} & \textbf{69.5}\sd{0.2} & \underline{70.2}\\
    \bottomrule
  \end{tabular}
\end{table*}

\subsection{Online streams}
\Cref{tab:online} reports the online protocol. The cache-based TDA and DMN are strongest on weakly correlated streams (67.7 and 67.2 at $\gamma{=}0.1$) but gain little on strongly correlated ones (66.8 and 66.3 at $\gamma{=}0.001$), whereas the anchored OGA, StatA and MOON improve with correlation (69.6, 69.5 and 69.9 at $\gamma{=}0.001$), since correlated batches contain few classes with many samples each. \method follows the trend of the anchored methods and adds 1.2, 1.3 and 1.1 points over StatA at medium and high correlation and on the separate stream, and 0.6 at low correlation (0.2 to 0.9 over MOON). The inductive variant, which predicts each batch before updating the statistics, retains most of the margin (69.5, 70.3 and 69.8 against 70.1, 70.8 and 70.2).

\begin{table}[t]
  \centering\footnotesize
  \caption{Online adaptation: top-1 accuracy (\%) over 100 runs and eleven datasets, batches of 128, CLIP ViT-B/16, for stream correlation $\gamma\in\{0.1,0.01,0.001\}$ and the separate-class stream.}
  \label{tab:online}
  \setlength{\tabcolsep}{4pt}
  \begin{tabular}{lcccc}
    \toprule
    \textbf{Method} & \textbf{Low} & \textbf{Medium} & \textbf{High} & \textbf{Separate}\\
    \midrule
    CLIP zero-shot~\cite{radford2021clip} & 65.2\sd{0.3} & 65.2\sd{0.3} & 65.2\sd{0.3} & 65.2\sd{0.3}\\
    TENT~\cite{wang2021tent} & 65.8\sd{0.4} & 65.5\sd{0.5} & 65.3\sd{0.6} & 64.5\sd{0.8}\\
    TDA~\cite{karmanov2024tda} & \textbf{67.7}\sd{0.4} & 67.1\sd{0.5} & 66.8\sd{0.6} & 66.6\sd{0.7}\\
    DMN~\cite{zhang2024dmn} & 67.2\sd{0.4} & 66.5\sd{0.5} & 66.3\sd{0.6} & 65.8\sd{0.7}\\
    OGA~\cite{fuchs2025oga} & 67.5\sd{0.5} & 69.1\sd{0.5} & 69.6\sd{0.6} & 69.1\sd{0.7}\\
    StatA~\cite{zanella2025stata} & 67.0\sd{0.5} & 68.9\sd{0.5} & 69.5\sd{0.6} & 69.1\sd{0.7}\\
    MOON~\cite{huang2026moon} & 67.4\sd{0.5} & 69.4\sd{0.5} & 69.9\sd{0.5} & 69.4\sd{0.6}\\
    \midrule
    \method (transductive) & \underline{67.6}\sd{0.5} & \textbf{70.1}\sd{0.4} & \textbf{70.8}\sd{0.5} & \textbf{70.2}\sd{0.6}\\
    \method (inductive) & 67.2\sd{0.5} & \underline{69.5}\sd{0.4} & \underline{70.3}\sd{0.5} & \underline{69.8}\sd{0.6}\\
    \bottomrule
  \end{tabular}
\end{table}

\subsection{Per-dataset results and distribution shifts}
\begin{figure}[t]
  \centering
  \includegraphics[width=0.75\linewidth]{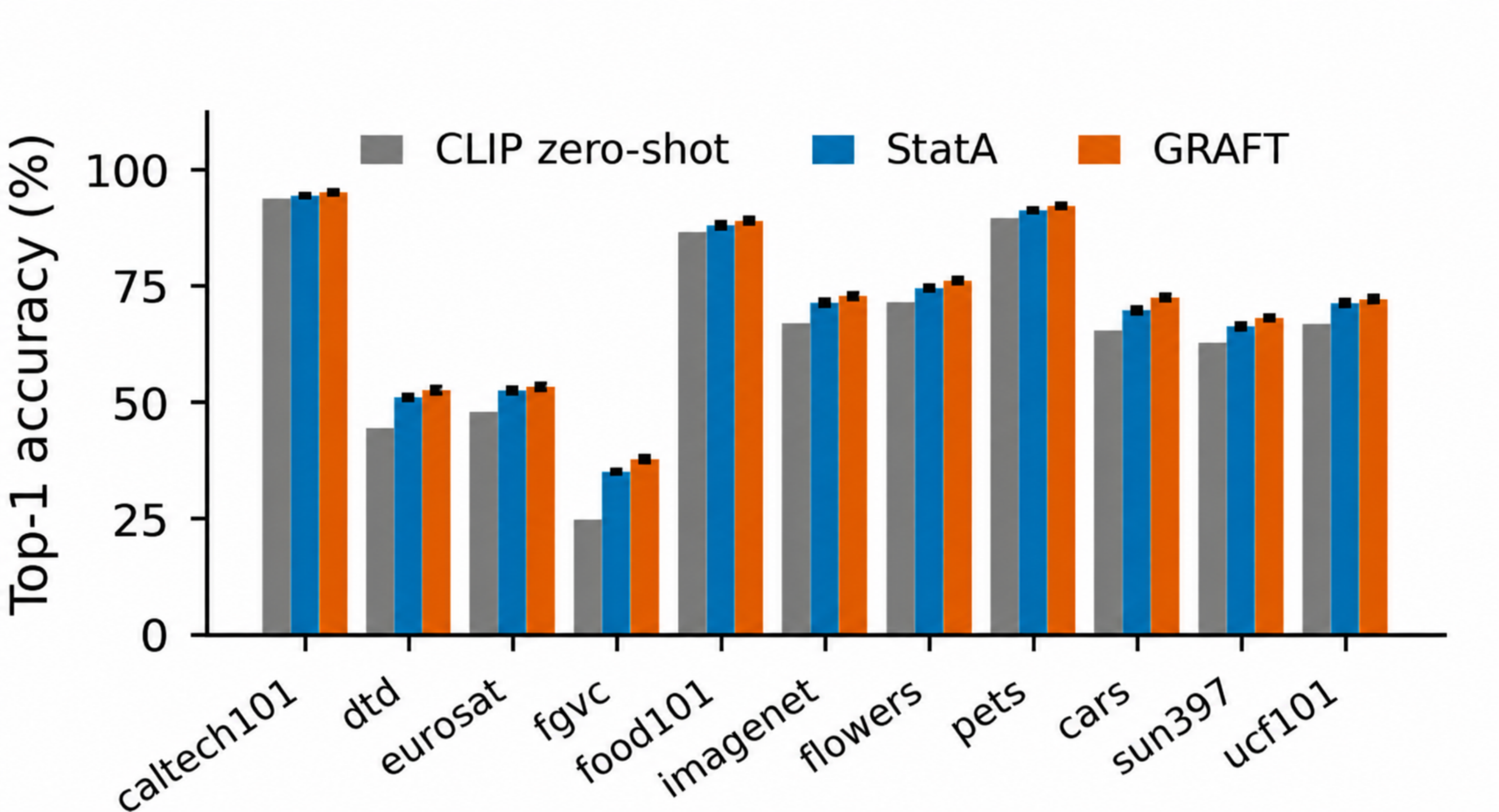}
  \caption{Per-dataset accuracy in the low setting for the zero-shot model, StatA and \method.}
  \label{fig:perdataset}
\end{figure}

\Cref{tab:perdataset} reports every baseline on each dataset in the low setting and \cref{fig:perdataset} plots the zero-shot model, StatA and \method. \method is first on eight of the eleven datasets and second on SUN397, EuroSAT and Flowers102, where EM-Dirichlet, strong at $B{=}64$ but collapsing with large batches, leads by 1.4, 4.3 and 0.4 points. Against StatA, \method improves on every dataset. The margin is largest on the fine-grained vocabularies with many confusable classes, Stanford Cars (2.8), FGVC-Aircraft (2.6) and SUN397 (1.9), and smallest on the coarse ones, Caltech101 (0.6), EuroSAT (0.7) and UCF101 (0.8). \Cref{tab:shift} extends the comparison to the ImageNet distribution shifts with the 1000-class vocabulary. The ranking is unchanged under the four shifts; StatA improves over zero-shot by 4.1 to 5.4 points, and the gain of \method over StatA grows from 1.3 points on ImageNet to 2.0 to 2.5 points on the shifted sets, where the displacement from the text anchors, and with it the shared component that the graph propagates, is largest.

\begin{table*}[t]
  \centering\scriptsize
  \caption{Per-dataset top-1 accuracy (\%) in the low setting ($B{=}64$, $K_{\mathrm{eff}}\in[2,10]$), CLIP ViT-B/16.}
  \label{tab:perdataset}
  \setlength{\tabcolsep}{3.6pt}
\begin{tabular}{lcccccccccccc}
\toprule
\textbf{Method} & \textbf{IN} & \textbf{SUN} & \textbf{Airc.} & \textbf{Euro.} & \textbf{Cars} & \textbf{Food} & \textbf{Pets} & \textbf{Flow.} & \textbf{Calt.} & \textbf{DTD} & \textbf{UCF} & \textbf{Avg.}\\
\midrule
CLIP zero-shot~\cite{radford2021clip} & 66.7 & 62.6 & 24.8 & 47.8 & 65.3 & 86.1 & 89.1 & 71.3 & 93.3 & 44.4 & 66.7 & 65.3\\
MTA~\cite{zanella2024mta} & 68.0 & 64.2 & 27.0 & 49.5 & 66.4 & 86.8 & 89.6 & 72.6 & 93.7 & 47.0 & 67.9 & 66.6\\
EM-Dirichlet~\cite{martin2024transductive} & 70.2 & \textbf{69.3} & 35.1 & \textbf{57.3} & 70.3 & \underline{88.0} & 91.0 & \textbf{76.1} & 94.1 & 50.6 & 71.4 & \underline{70.3}\\
ZLaP~\cite{kalantidis2024zlap} & 35.4 & 30.7 & 13.3 & 28.4 & 40.3 & 48.4 & 47.7 & 38.2 & 44.9 & 26.6 & 33.3 & 35.2\\
TransCLIP~\cite{zanella2024transclip} & 43.2 & 35.8 & 16.8 & 32.6 & 44.4 & 50.8 & 48.8 & 45.9 & 62.5 & 24.2 & 39.5 & 40.4\\
StatA~\cite{zanella2025stata} & 71.0 & 66.0 & 35.0 & 52.3 & 69.6 & 87.4 & 90.7 & 74.2 & 94.0 & 50.9 & 71.1 & 69.3\\
MOON~\cite{huang2026moon} & \underline{71.7} & 66.9 & \underline{36.3} & 52.6 & \underline{71.0} & 87.9 & \underline{91.2} & 75.0 & \underline{94.3} & \underline{51.6} & \underline{71.5} & 70.0\\
\midrule
\method & \textbf{72.3} & \underline{67.9} & \textbf{37.6} & \underline{53.0} & \textbf{72.4} & \textbf{88.5} & \textbf{91.6} & \underline{75.7} & \textbf{94.6} & \textbf{52.3} & \textbf{71.9} & \textbf{70.7}\\
\bottomrule
\end{tabular}

\end{table*}

\begin{figure}[t]
  \centering
  \includegraphics[width=0.62\linewidth]{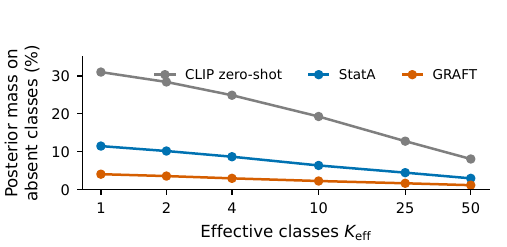}
  \caption{Posterior mass assigned to classes absent from the batch versus the number of effective classes (ImageNet, $B{=}64$).}
    \label{fig:absent}
\end{figure}

\begin{table}[t]
  \centering\scriptsize
  \caption{ImageNet (IN) and its shifts V2, Sketch (S), A and R in the low setting, top-1 accuracy (\%), CLIP ViT-B/16.}
  \label{tab:shift}
  \setlength{\tabcolsep}{3.6pt}
\begin{tabular}{lccccc}
\toprule
\textbf{Method} & \textbf{IN} & \textbf{V2} & \textbf{S} & \textbf{A} & \textbf{R}\\
\midrule
CLIP zero-shot & 66.7 & 60.8 & 46.2 & 47.8 & 74.0\\
TransCLIP & 43.2 & 35.5 & 29.1 & 28.9 & 42.8\\
StatA & 71.0 & 66.2 & 51.1 & 51.9 & 79.4\\
MOON & \underline{71.7} & \underline{66.8} & \underline{51.7} & \underline{52.6} & \underline{79.7}\\
\method & \textbf{72.3} & \textbf{68.3} & \textbf{53.1} & \textbf{54.3} & \textbf{81.9}\\
\bottomrule
\end{tabular}
\end{table}

\subsection{Qualitative results}
\begin{figure}[t]
  \centering
  \includegraphics[width=0.78\linewidth]{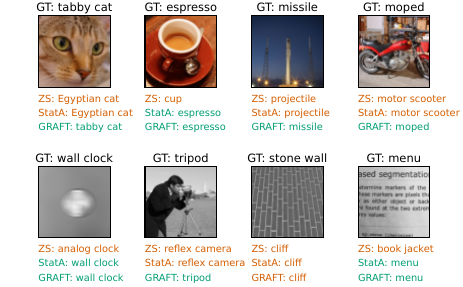}
  \caption{Qualitative predictions in the low setting: ground truth (GT) and the predictions of the zero-shot model (ZS), StatA and \method. Correct predictions in green, wrong ones in red.}
  \label{fig:qualgrid}
\end{figure}

\Cref{fig:qualgrid} shows test images from batches of the low setting with the ground truth and the predictions of the zero-shot model, StatA and \method. The typical correction is a sample of a present class that the zero-shot model assigns to a semantically close class absent from the batch (an Egyptian cat for a tabby cat). StatA recovers some of these samples when the present class has enough evidence to move its prototype, and \method recovers the remaining ones because the absent neighbour is removed through the presence model and displaced consistently with the present class. The failure case, a stone wall predicted as a cliff by all three methods, has no close neighbour in the batch. \Cref{fig:absent} quantifies the mechanism on ImageNet batches of 64: with a single present class the zero-shot model assigns 31.0\% of its posterior mass to classes that are not in the batch, StatA 11.5\% and \method 4.1\%. With fifty present classes the values converge to 8.1\%, 3.0\% and 1.2\%.

\section{Ablation Studies}
\label{sec:ablation}

\paragraph{Components.}
\Cref{tab:components} adds the components one at a time. Without anchor, coupling or presence the mixture reaches 43.3 in the low setting, 22 points below the zero-shot model, but 70.3 on the full set, where every class is well observed. The presence model alone raises the low-setting result to 55.8, the anchor term alone to 69.1, and the two together to 69.6, the anchor-only regime of StatA and MOON. The graph coupling adds 1.1 points on top (70.7), the margin that distinguishes \method from the anchored methods, and empirical Bayes a further 0.1. The inherited sample graph contributes 0.3 points in the low setting but 1.5 on the full set, where the anchor and the coupling change the result by at most 0.4 points.

\begin{table}[t]
  \centering\scriptsize
  \caption{Component ablation, top-1 accuracy (\%): $\pi$ presence, $\alpha$ anchor, $\lambda$ graph coupling, EB empirical Bayes, $W_s$ sample graph.}
  \label{tab:components}
  \setlength{\tabcolsep}{4pt}
  \begin{tabular}{ccccc@{\hskip 8pt}cc}
    \toprule
    $\boldsymbol{\pi}$ & $\boldsymbol{\alpha}$ & $\boldsymbol{\lambda}$ & \textbf{EB} & $\boldsymbol{W_s}$ & \textbf{Low} ($\boldsymbol{B=64}$) & \textbf{All}\\
    \midrule
    \multicolumn{5}{c}{zero-shot} & 65.2 & 65.2\\
    -- & -- & -- & -- & \checkmark & 43.3\sd{0.5} & 70.3\\
    \checkmark & -- & -- & -- & \checkmark & 55.8\sd{0.5} & 70.3\\
    -- & \checkmark & -- & -- & \checkmark & 69.1\sd{0.3} & 69.9\\
    \checkmark & \checkmark & -- & -- & \checkmark & 69.6\sd{0.3} & 69.9\\
    -- & \checkmark & \checkmark & -- & \checkmark & 70.2\sd{0.3} & 70.2\\
    \checkmark & \checkmark & \checkmark & -- & -- & 70.4\sd{0.3} & 68.7\\
    \checkmark & \checkmark & \checkmark & -- & \checkmark & 70.7\sd{0.3} & 70.2\\
    \checkmark & \checkmark & \checkmark & \checkmark & \checkmark & \textbf{70.8}\sd{0.3} & 70.2\\
    \bottomrule
  \end{tabular}
\end{table}

\paragraph{Prior and graph.}
We vary $\alpha$, $\lambda$ and the number of neighbours $m$, and replace the text kNN graph by a complete graph and by a degree-matched random graph (table and sensitivity curves in the supplementary material). Accuracy in the low setting varies by at most 0.6 points for $\lambda$ between 1 and 16 and by 0.3 points for $m$ between 3 and 10, so neither is a sensitive hyper-parameter; removing the anchor ($\alpha=0$) costs 1.7 points and removing the coupling ($\lambda=0$) 1.6. The random graph (69.2) performs like no coupling (69.1) and the complete graph (68.8) is below both, so pooling helps only when it follows the semantic neighbourhoods, the control that H1 of \cref{sec:field} requires. H1 itself is tested with labels (supplementary material): the energy ratio of the oracle displacement field on the text graph against degree-matched random graphs lies between 0.25 and 0.75 on all datasets, and the datasets with the highest ratios have the smallest margins in \cref{tab:perdataset}. Likelihood and optimisation variants change the result by at most 0.5 points.

\section{Conclusion}
\label{sec:conclusion}
We traced the collapse of unlabeled VLM adaptation on realistic batches to independent per-class parameters and replaced them by a displacement field over the vocabulary with a Gaussian Markov random field prior on a semantic class graph. \method is closed form, contains anchor-only shrinkage and a global shift as extreme settings, and improves over anchored transduction on realistic batches.

\section*{LLM Usage Statement}
Generative AI tools were used for language editing of the manuscript and to assist with the literature search. They were not used to generate experimental results, numerical values, or the statements and proofs of the theoretical results, all of which were derived by the authors. The authors take full responsibility for the final content of this work, including all text, claims and artifacts.

{
    \small
    \bibliographystyle{ieeenat_fullname}
    \bibliography{references}
}

\end{document}